\documentclass{svproc}
\usepackage{url}

\usepackage{times}
\usepackage{epsfig}
\usepackage{graphicx}
\usepackage{amsmath}
\usepackage{amssymb}
\usepackage{csquotes}
\usepackage{algorithm}
\usepackage{algorithmicx}
\usepackage{algpseudocode}
\usepackage{mathtools}
\usepackage{amsfonts}
\usepackage{csquotes}

\algdef{SE}[DOWHILE]{Do}{doWhile}{\algorithmicdo}[1]{\algorithmicwhile\ #1}
\usepackage{hyperref}
\usepackage{tikz}

\begin{document}
\mainmatter              
\title{Deep Convolutional Likelihood Particle Filter for Visual Tracking}
\titlerunning{Likelihood Particle Filter for Visual Tracking}  
%
\author{Reza Jalil Mozhdehi and Henry Medeiros}
\authorrunning{Mozhdehi and Medeiros} 
%
%
\institute{Marquette University, Milwaukee, WI, USA,\\
\email{reza.jalilmozhdehi@marquette.edu} and \email{henry.medeiros@marquette.edu}}

\maketitle              

%

\begin{abstract}
We propose a novel particle filter for convolutional-correlation visual trackers. Our method uses correlation response maps to estimate likelihood distributions and employs these likelihoods as proposal densities to sample particles. Likelihood distributions are more reliable than proposal densities based on target transition distributions because correlation response maps provide additional information regarding the target's location. Additionally, our particle filter searches for multiple modes in the likelihood distribution, which improves performance in target occlusion scenarios while decreasing computational costs by more efficiently sampling particles. In other challenging scenarios such as those involving motion blur, where only one mode is present but a larger search area may be necessary, our particle filter allows for the variance of the likelihood distribution to increase. We tested our algorithm on the Visual Tracker Benchmark v1.1 (OTB100) and our experimental results demonstrate that our framework outperforms state-of-the-art methods.
\end{abstract}

\section{Introduction}
\label{sec:1} 

Particle filters are widely applied in visual tracking problems due to their ability to find targets in challenging scenarios such as those involving occlusions or fast motion. Recently, particle filters have been used in conjunction with deep convolutional neural networks (CNN) \cite{NIPS2012,234} and correlation filters \cite{dai2019visual,zhang2018visual,qi2016hedged,77}. The Hierarchical Convolutional Feature Tracker (HCFT) proposed by Ma et al. in \cite{77} showed significant performance improvements over previous works, demonstrating the effectiveness of using convolutional features along with correlation filters. Correlation filters provide a map showing similarities between convolutional features corresponding to an image patch and the target \cite{dai2019visual,zhang2018visual,777}. Adding a particle filter to convolutional-correlation visual trackers can significantly improve their results as shown in \cite{CPF,Redetection,mozhdehideep,mozhdehideep2,mozhdehideep3}. In these methods, particle filters sample several image patches and calculate the weight of each sample by applying a correlation filter to the convolutional response maps.

In this work, we propose a novel convolutional-correlation particle filter for visual tracking which estimates likelihood distributions from correlation response maps. Sampling particles from likelihood distributions improves the accuracy of patch candidates because correlation response maps have an initial evaluation of the target location. Thus, they are more reliable proposal densities than transition distributions, commonly used  in particle-correlation trackers such as \cite{CPF,Redetection,mozhdehideep,mozhdehideep2}. Additionally, these trackers calculate the posterior distribution based on the peaks of correlation maps without considering them in the computation of particle weights. Our particle filter solves this problem using a multi-modal likelihood distribution to address challenging tracking scenarios. Our proposed algorithm also calculates a likelihood distribution with larger variances, which is useful in other challenging scenarios involving fast motion or background clutter because it expands the target search area. Additionally, this method decreases the number of required particles. Experimental results on the Visual Tracker Benchmark v1.1 (OTB100) \cite{WuLimYang} show that our proposed framework outperforms state-of-the-art methods.

\section{The change of support problem in convolution-correlation particle filters}
\label{sec:2} 

The particle weights in a particle filter are calculated by \cite{tutorial}
\begin{equation} \label{eq:w_xt}
     \omega^{(i)}_{x_{t}} \propto \omega^{(i)}_{x_{t-1}} \dfrac {p(y_{t}|x^{(i)}_{t}) p(x^{(i)}_{t}|x_{t-1})}{q(x^{(i)}_{t}|x_{t-1},y_{t})},
\end{equation}
where $p(x^{(i)}_{t}|x_{t-1})$ and $p(y_{t}|x^{(i)}_{t})$ are the transition and likelihood distributions, and  $q(x^{(i)}_{t}|x_{t-1},y_{t})$ is the proposal distribution used to sample the particles. The posterior distribution is then approximated by
\begin{equation} \label{eq:2}
\hat{Pr}(x_{t}|y_{t}) \approx \sum_{i=1}^{N} \varpi^{(i)}_{x_{t}} \delta(x_{t}-x^{(i)}_{t}),
\end{equation}
where $\varpi^{(i)}_{t}$ are the normalized weights. However,  particle filters used in correlation trackers generally sample particles from the transition distribution, i.e., $q(x^{(i)}_{t}|x_{t-1},y_{t})=p(x^{(i)}_{t}|x_{t-1})$. These methods also re-sample particles at every frame, which  removes the term corresponding to previous weights $\omega^{(i)}_{x_{t-1}}$ from Eq. \ref{eq:w_xt}. Finally, the weight of each particle in these trackers is given by \cite{CPF}
\begin{equation} \label{eq:3}
     \omega^{(i)}_{x_{t}} \propto p(y_{t}|x^{(i)}_{t}),
\end{equation}
where $p(y_{t}|x^{(i)}_{t})$ is a function of $R_{x_{t}^{(i)}}^{y_{t}} \in \mathbb{R}^{M \times Q}$, the correlation response map centered at $x_{t}^{(i)}$. In these trackers, particles are shifted to the peaks of correlation maps and the posterior distribution is then approximated by the particles' weights at the shifted locations, i.e., \begin{equation} \label{eq:4}
\hat{Pr}(x_{t}|y_{t}) \approx \sum_{i=1}^{N} \varpi^{(i)}_{x_{t}} \delta(x_{t}-\tilde{x}^{(i)}_{t}),
\end{equation}
where $\tilde{x}^{(i)}_{t}$ is the peak of the correlation response map corresponding to the $i$-th particle. However, the posterior distribution using the shifted locations must consider the weights corresponding to the new support points, not the original locations of the particles. That is, the original locations are used in weight computation, but the shifted support is used to approximate the posterior distribution. To solve this, we sample particles from the likelihood distribution instead. Particle filters that sample from likelihood distributions generate more accurate particles, but sampling from the likelihood distribution is not always possible. Fortunately, convolutional-correlation trackers generate correlation maps that can be used in the construction of likelihood distributions. 

\section{Likelihood Particle Filter}
\label{sec:3}

\begin{figure}[t]
\centering
    \includegraphics[width=.9\textwidth]{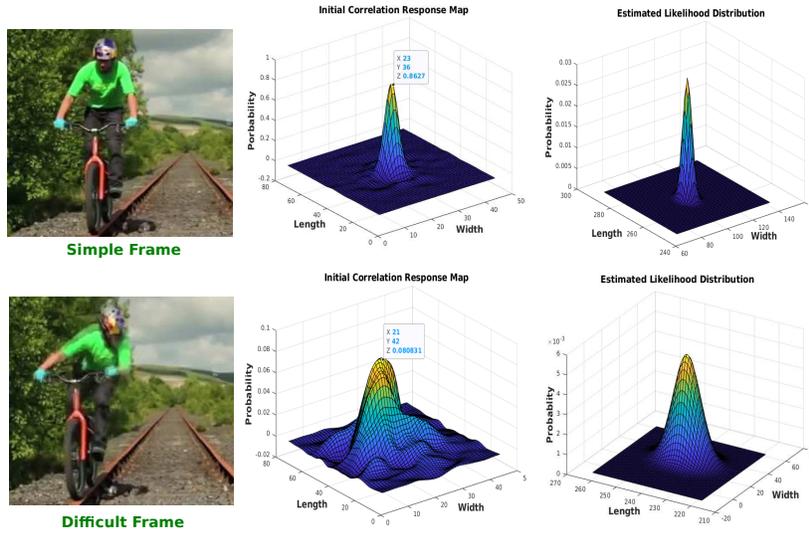}
\caption{Estimated likelihood distributions for common scenarios (simple frame) and a challenging scenario involving fast motion (difficult frame).}
\label{fig:1new}
\end{figure}

Our algorithm generates an initial correlation response map for the current frame based on the previously estimated target state to calculate an initial likelihood distribution. That is, we generate a patch from the current frame based on the previous target state and use a CNN \cite{234} to extract the convolutional features from this patch. We then compare these features with the target model to calculate the final correlation response map \cite{77}. 
As seen in Fig. \ref{fig:1new}, in most scenarios (which we call ``simple frames'') the correlation response map corresponds to a sharp Gaussian distribution with a prominent peak. In challenging scenarios (``difficult frames''), correlation maps are wider with less pronounced peaks. We need to estimate likelihood distributions consistently in both scenarios. To address this issue, we fit a Gaussian distribution to the correlation response maps while disregarding elements with probability lower than a threshold $\tau$. By disregarding low probability elements, we mitigate the impact of the background on the computation of the model. We compute the mean of the correlation response map using
\begin{equation} \label{eq:5}
\mu \approx \dfrac{\sum_{i=1}^{u} q_{i} s_{i}}{\sum_{i=1}^{u} q_{i}},
\end{equation}
where $s_{i}$ and $q_{i}$ represent the elements of the correlation response map and their respective probabilities, and $u$ is the number of elements with probability higher than $\tau$. The variance of the response map is then given by
\begin{equation} \label{eq:6}
\sigma^{2} \approx \dfrac{\sum_{i=1}^{u} q_{i} (s_{i} - \mu)^{2}}{\sum_{i=1}^{u} q{i}}.
\end{equation}
Thus, our model assigns low probabilities to pixels that are likely to belong to the background while assigning relatively high probabilities to all the regions that might correspond to the target. As a result, our samples concentrate in regions where the target is more likely to be present.

\begin{figure}
\centering
    \includegraphics[width=.6\textwidth]{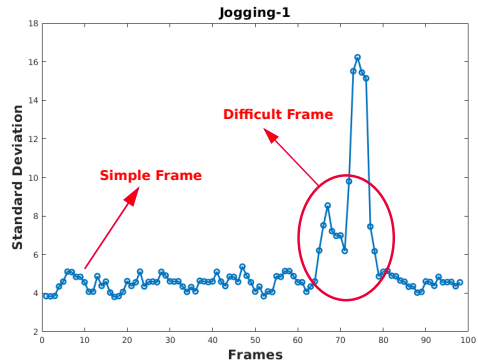}
\caption{Standard deviations of the estimated likelihood distributions in data sequence \emph{Jogging-1} of the OTB-100 dataset.}
\label{fig:3}
\end{figure}

Fig. \ref{fig:1new} shows our estimated likelihood distributions for two different frames of the \emph{Biker} data sequence of the OTB100 benchmark. In the difficult frame, the target undergoes motion blur, which causes the correlation response map to be wider with a lower peak. Our estimated variance is then correspondingly higher, which helps our tracker to sample particles over a wider area to compensate for tracking uncertainties in difficult scenarios. The example in Fig. \ref{fig:3} shows how the variance increases as the target approaches difficult frames. 

Although allowing for higher variances in challenging scenarios such as those involving fast motion helps our tracker address such issues, this strategy alone cannot handle multi-modal correlation response maps. To resolve this issue, we propose to determine the peaks of the distribution using the approach described below.

\subsection{Multi-modal likelihood estimation}

\begin{figure}
\centering
    \includegraphics[width=.99\textwidth]{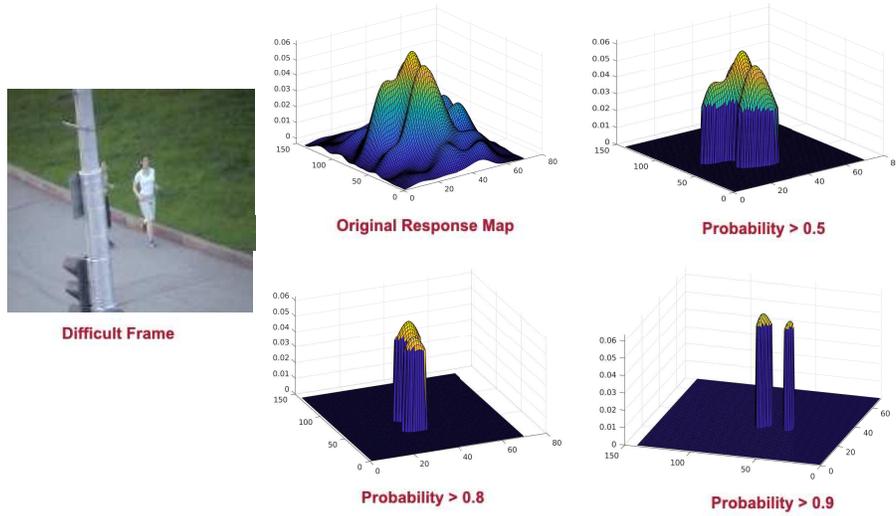}
\caption{A difficult frame including target occlusion. Its correlation response map has two peaks. By increasing the threshold to remove low probability elements, two clusters corresponding to the target and the pole are seen.}
\label{fig:4}
\end{figure}

The existence of multiple peaks in a correlation response map usually indicates the presence of confusing elements in the background of the frame, as the example in Fig. \ref{fig:4} illustrates. In the frame shown in the figure, there are two peaks in the correlation response map when partial target occlusion occurs. The peaks correspond to the woman on the left side of the image (the target) and the pole partially occluding her. By applying a threshold to remove low probability elements from the correlation response map, two clusters become apparent. 

\begin{figure}
\centering
    \includegraphics[width=.99\textwidth]{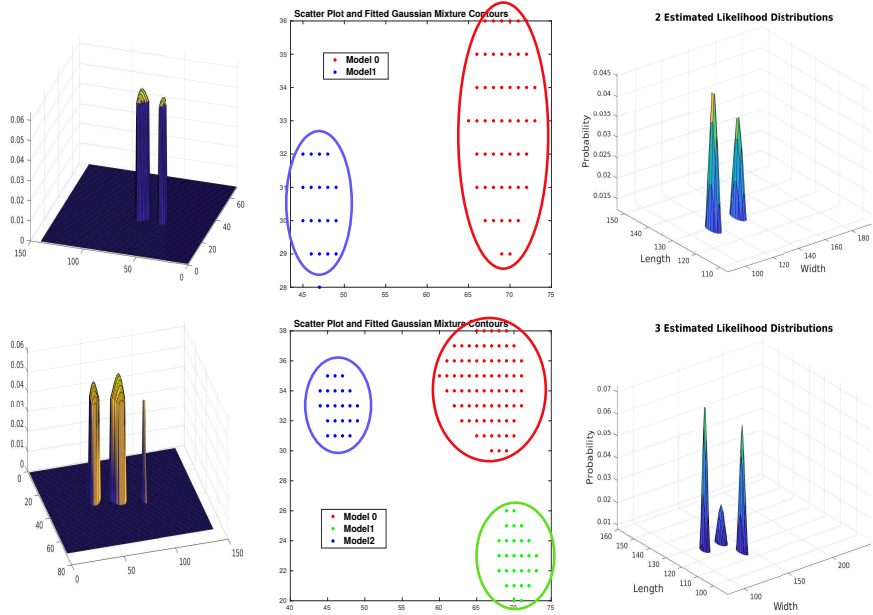}
\caption{Finding clusters; left: correlation response maps with two and three clusters, middle: clusters of the correlation response maps obtained by fitting a Gaussian mixture model, right: estimated likelihood distributions for each cluster.}
\label{fig:finding_clusters}
\end{figure}

To identify the peaks of the correlation map while disregarding additional background clutter, we remove from the map points with probability lower than a threshold $\tau$. We then  fit a Gaussian mixture model to the remaining feature map points which clusters them into $k$  groups \cite{mixture}. Fig. \ref{fig:finding_clusters} shows two instances of correlation response maps in which we identify $k=2$ and $k=3$ clusters. The likelihood corresponding to each peak is then given by a normal distribution with mean and variance given by Eqs. \ref{eq:5} and \ref{eq:6}.  Algorithm \ref{alg:1} summarizes our proposed approach to estimate the likelihood distribution for each cluster.

\begin{algorithm}
\caption{Multi-modal likelihood estimation.} 
\label{alg:1}
     \begin{algorithmic}[1]
         \Require{Current frame $y_t$ and previous target state $x_{t-1}$}
         \Ensure{One likelihood distribution for each correlation map cluster}
         \State {Extract a patch from the current frame based on the previous target state}
         \State {Extract the CNN features of the patch and calculate its correlation response map}
         \State {Remove points with probability lower than $\tau$}
         \State {Fit a Gaussian mixture model to the map and find the clusters}
         \State {Estimate the likelihood distribution of each cluster based on the mean and variance of its elements in the map according to Eqs. \ref{eq:5} and \ref{eq:6}}
     \end{algorithmic}
\end{algorithm}

\subsection{Particle sampling}
We sample particles from the Gaussian likelihood distributions obtained from the correlation response maps in the current frame. The probability that a particle is sampled from the likelihood distribution is given by
\begin{equation}
p(x_{t}^{(i)} \vert y_{t})\propto \sum_{j=1}^{k}{\mathcal{N}\left(x_t^{(i)};\mu_j, \sigma_j \right)},  \label{eq:p_xt} \end{equation}
where $\mu_j$ and $\sigma_j$ are the mean and variance of the $j$-th mode of the likelihood. We generate a patch for each particle and extract its features using a CNN. After calculating the correlation response map for each particle, we shift the particles to the peaks of their respective correlation response maps. The peak of each correlation response map is the estimated target position based on the patch centered at the corresponding particle. Because each particle is shifted to the peak of the correlation response map, we consider $p(\tilde{x}_{t}^{(i)} \vert x_{t}^{(i)}) = 1$, where $\tilde{x}_{t}^{(i)}$ is the peak of the corresponding correlation response map. As a result, $p(x_{t}^{(i)}|y_{t})=p(\tilde{x}_{t}^{(i)}|y_{t})$.

\begin{figure}[t]
\centering
    \includegraphics[width=.99\textwidth]{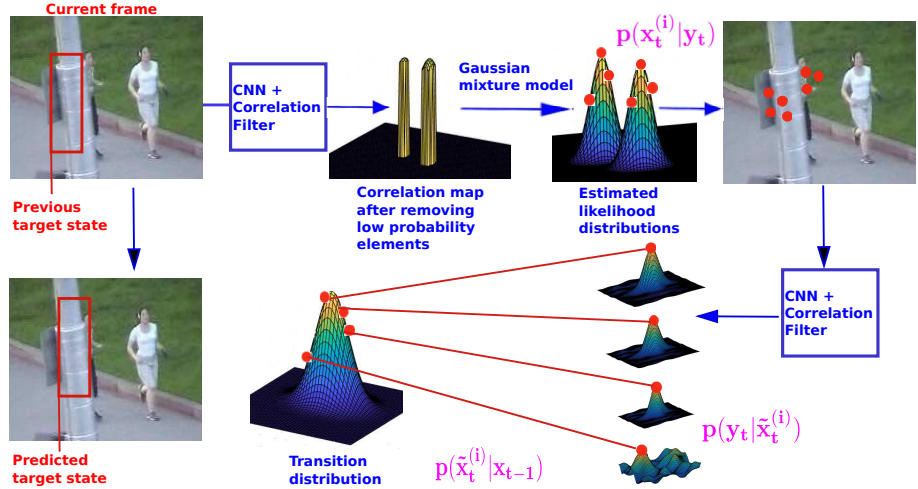}
\caption{Overview of the steps comprising the proposed DCPF-Likelihood visual tracker.}
\label{fig:6}
\end{figure}

\subsection{Calculating the weights and posterior distribution}

By computing the weight of each shifted particle $\tilde{x}_{t}^{(i)}$, we can accurately estimate the posterior based on the shifted particles and their correct weights, which addresses the problem of incorrect support points observed in previous works. As discussed earlier, Eq. \ref{eq:w_xt} corresponds to the weight of each particle before shifting. The weight of the shifted particles is then given by 
\begin{equation} \label{eq:7}
     \omega^{(i)}_{\tilde{x}_{t}} \propto \omega^{(i)}_{x_{t-1}} \dfrac {p(y_{t}|\tilde{x}^{(i)}_{t}) p(\tilde{x}^{(i)}_{t}|x_{t-1})}{q(\tilde{x}^{(i)}_{t}|x_{t-1},y_{t})},
\end{equation}
where the term corresponding to the previous weight is removed because we perform resampling at every frame. Additionally, \cite{tutorial}
\begin{equation}
  q(\tilde{x}^{(i)}_{t}|x_{t-1},y_{t}) =  p(\tilde{x}_{t}^{(i)} \vert y_{t}). \label{eq:q}
\end{equation}
Thus, the weight of each shifted particle is
\begin{equation} \label{eq:9}
     \omega^{(i)}_{\tilde{x}_{t}} \propto  \dfrac {p(y_{t}|\tilde{x}^{(i)}_{t}) p(\tilde{x}^{(i)}_{t}|x_{t-1})}{p(\tilde{x}_{t}^{(i)} \vert y_{t})}.
\end{equation}
Let the target state be defined as
\begin{equation}\label{eq:10}
    z_{t-1} = \begin{bmatrix}
    x_{t-1},
    \dot{x}_{t-1}
    \end{bmatrix}^T,
\end{equation}
where $\dot{x}_{t-1}$ is the velocity of $x_{t-1}$. We apply a first-order motion model to $z_{t-1}$ according to
\begin{equation} \label{eq:11}
  \bar{z}_{t-1} = A z_{t-1},
\end{equation}
where where $\bar{z}_{t-1}$ represents the predicted target state and $A$ is the process matrix defined by 
\begin{equation} \label{eq:12}
   A = \left[ \begin{array}{c|c}
    I_4  & I_4 \\ \hline
    0_{(4,4)} & I_4 
\end{array} \right],
\end{equation}
where $I_4$ is a $4 \times 4$ identity matrix and $0_{(4,4)}$ is a $4 \times 4$ zero matrix. We use a Gaussian distribution $\mathcal{N}(\bar{x}_{t-1},\sigma^{2})$ to find the probability of each estimated particle in the current frame $p(\tilde{x}^{(i)}_{t}|x_{t-1})$.

Additionally, $p(y_{t}|\tilde{x}^{(i)}_{t})$ is the likelihood of each shifted particle. Let $f_{x_{t}^{(i)}}(l,o)$ be the convolutional features of each particle $x_{t}^{(i)}$ where $l$ and $o$ represent the layers and channels of the network, respectively. The correlation response map is then calculated by \cite{77}
\begin{equation}\label{eq:13}
R_{x_{t}^{(i)}}^{y_{t}}(x) = \sum_{l=1}^{L} \Upsilon_l (\mathfrak{F}^{-1}(\sum_{o=1}^{O} C_{t-1}(l,o)\odot\bar{F}_{x_{t}^{(i)}}(l,o))),
\end{equation}
where $\bar{F}_{x_{t}^{(i)}}(l,o)$ is the complex conjugate Fourier transform of $f_{x_{t}^{(i)}}(l,o)$, $C_{t-1}$ is the model generated in the previous frame, $\odot$ represents the Hadamard product,  $\mathfrak{F}^-1$ is the inverse Fourier transform operator, and $\Upsilon_l$ is a regularization term \cite{77}.
The peak of $R_{x_{t}^{(i)}}^{y_{t}}$ is then calculated by
\begin{equation} \label{eq:14}
   \tilde{x}^{(i)}_{t} =\arg\max_{m,q}R_{x_{t}^{(i)}}^{y_{t}}(m,q),
\end{equation}
where $m = 1,..., M$ and $q = 1,..., Q$. The likelihood of $\tilde{x}^{(i)}_{t}$ is calculated by \cite{mozhdehideep2}
\begin{equation} \label{eq:15}
p(y_{t}|\tilde{x}^{(i)}_{t})=\dfrac{1}{M \times Q}\sum_{m,q}R_{\tilde{x}_{t}^{(i)}}^{y_{t}}(m,q).
\end{equation}
The posterior distribution based on the shifted particles and their respective weights is then 
\begin{equation} \label{eq:16}
\hat{Pr}(x_{t}|y_{t}) \approx \sum_{i=1}^{N} \varpi^{(i)}_{\tilde{x}_{t}} \delta(x_{t}-\tilde{x}^{(i)}_{t}),
\end{equation}
where $\varpi^{(i)}_{\tilde{x}_{t}}$ is the normalized version of $\omega^{(i)}_{\tilde{x}_{t}}$. Fig. \ref{fig:6} summarizes the steps of our method, and Algorithm \ref{alg:2} describes the details of our approach.
\begin{algorithm}
\caption{DCPF-Likelihood visual tracker.} 
\label{alg:2}
     \begin{algorithmic}[1]
         \Require{Current frame $y_t$ and previous target state $x_{t-1}$}
         \Ensure{Current target state $x_{t}$}
         \State {Estimate a likelihood distribution for each cluster using Algorithm \ref{alg:1}}
         \State {Sample particles from the likelihood distributions $p(x_{t}^{(i)} \vert y_{t})$}
         \State {Extract the CNN features of the patches corresponding to each particle and calculate its correlation response map according to Eq. \ref{eq:13}}
         \State {Shift the particles to the peaks of their correlation response maps based on Eq. \ref{eq:14}}
         \State {Calculate the likelihood $p(y_{t}|\tilde{x}^{(i)}_{t})$ based on Eq. \ref{eq:15}}
         \State {Calculate the transition probability $p(\tilde{x}^{(i)}_{t}|x_{t-1})$ according to Eqs. \ref{eq:10} to \ref{eq:12}}
         \State {Compute the weight of each shifted particle $\omega^{(i)}_{\tilde{x}_{t}}$ according to Eqs. \ref{eq:7} to \ref{eq:9}}
         \State {Calculate the posterior distribution according to Eq. \ref{eq:16}}
     \end{algorithmic}
\end{algorithm}

\section{Experimental results}
\label{subsec:2}

\begin{figure}[t]
\centering
    \includegraphics[width=.99\textwidth]{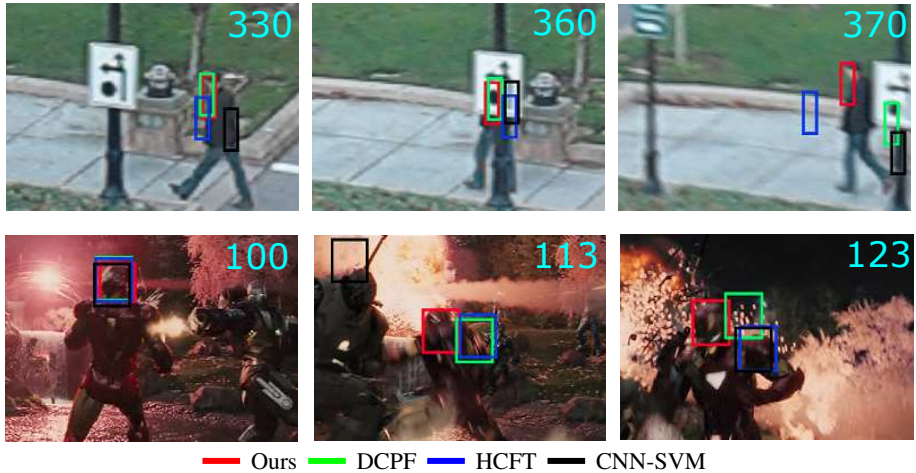}
    \begin{tikzpicture}
 \draw [line width=0.8mm, red,left] (0,0) -- (.5,0) node 
 [right,color=black] (text1) {Ours};;
 \draw [line width=0.8mm, green] (text1.east) -- 
 ([xshift=5mm]text1.east) node [right,color=black] (text2) {DCPF};;
 \draw [line width=0.8mm, blue] (text2.east) -- 
 ([xshift=5mm]text2.east) node [right,color=black] (text3) {HCFT};;
 \draw [line width=0.8mm, black] (text3.east) -- 
 ([xshift=5mm]text3.east) node [right,color=black] (text4) {CNN-SVM};;
\end{tikzpicture}
\begin{small}
\caption{Qualitative evaluation of our tracker against \textit{DCPF}, \textit{HCFT}, and \textit{CNN-SVM} on two challenging sequences: \textit{Human6} (top) and \textit{Ironman} (bottom).}
\label{fig:OPE}
\end{small}
\end{figure}

We use the Visual Tracker Benchmark v1.1 (OTB100) to assess the performance of our tracker. This benchmark contains 100 video sequences, which include 11 challenging scenarios. Our results are based on the one-pass evaluation (OPE), which uses the ground truth target size and position in the first frame to initialize the tracker. Our evaluation is based on the precision and success measures, described in \cite{WuLimYang}. 
Fig. \ref{fig:OPE} shows qualitative results comparing our tracker with DCPF \cite{mozhdehideep}, HCFT \cite{77}, and CNN-SVM \cite{hong2015tracking}. In both data sequences shown in the figure, our method successfully handles occlusion scenarios. These results highlight the impact of using more reliable sampling distributions.

\begin{figure*}[t]
     \centering 
\begin{minipage}{1\linewidth}
\centering
\includegraphics[width=.32\textwidth]{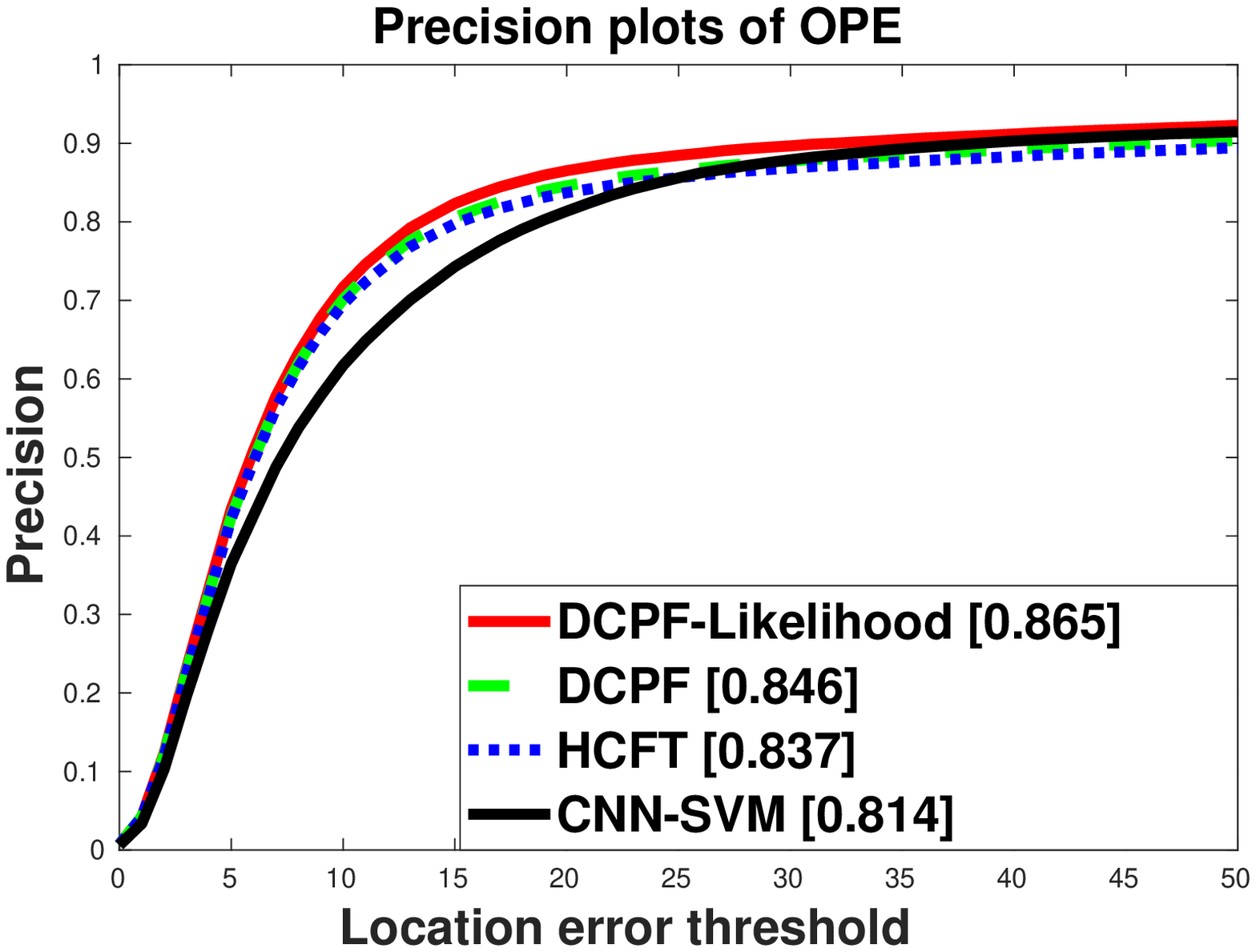}
\includegraphics[width=.32\textwidth]{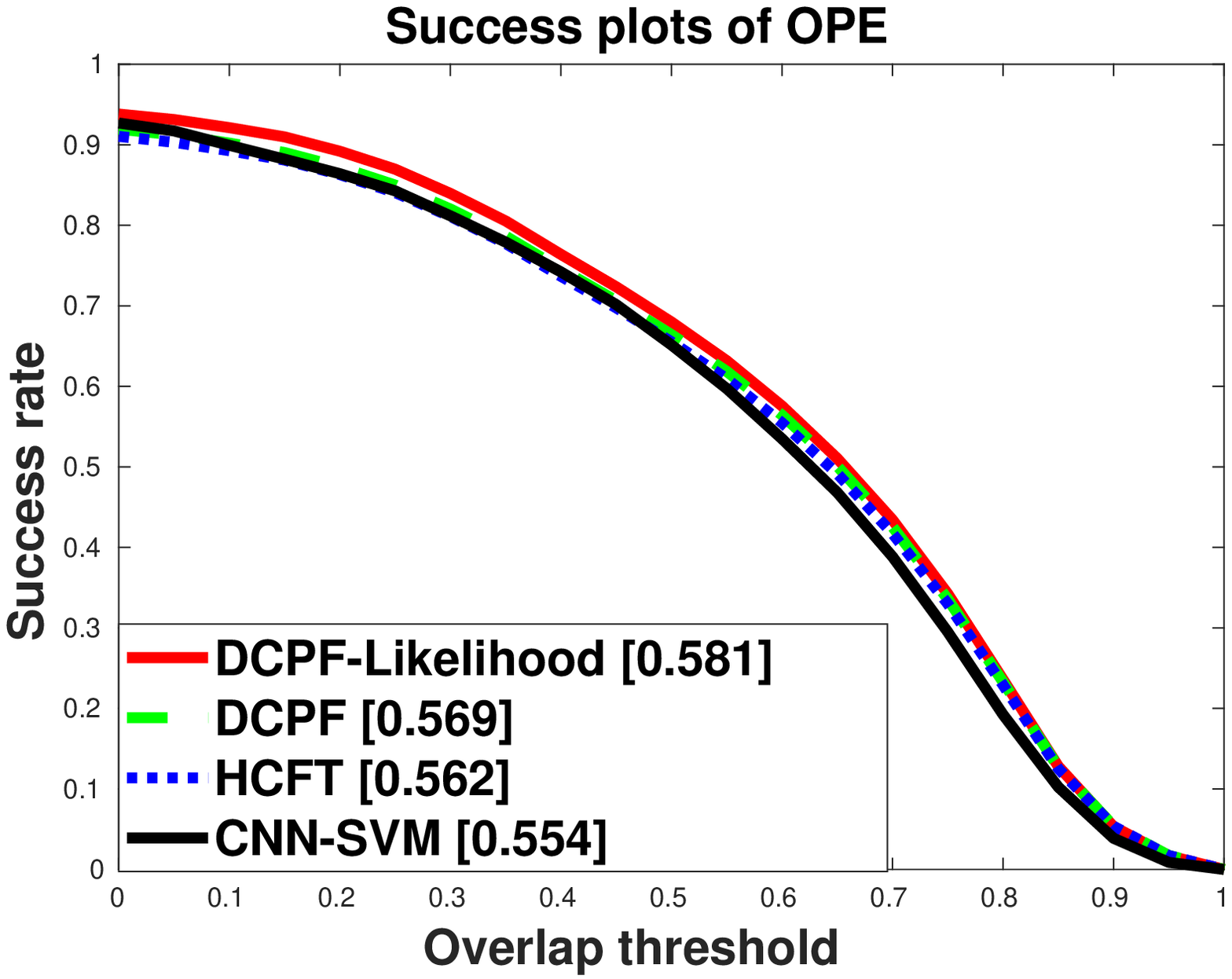}
\includegraphics[width=.32\textwidth]{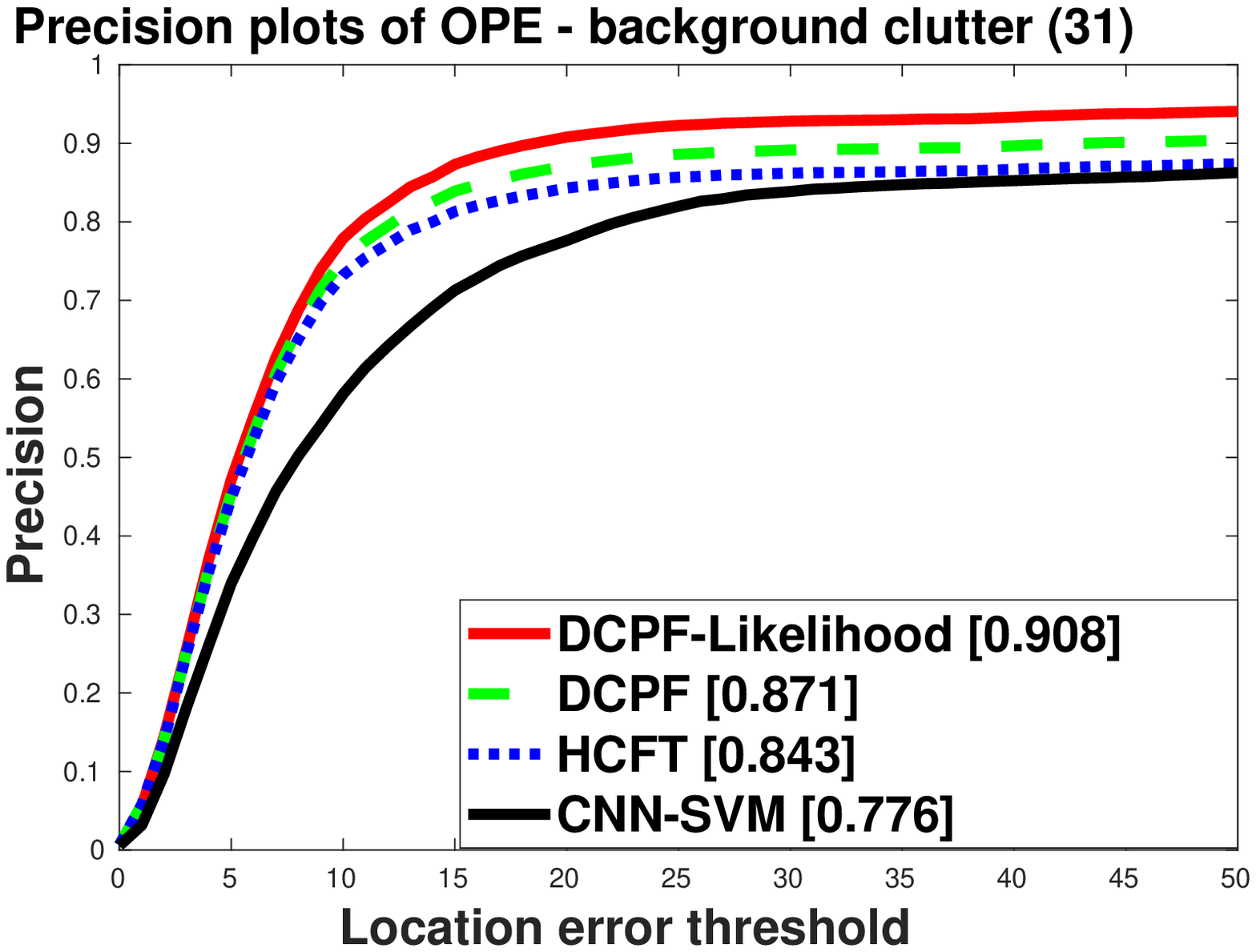}
\end{minipage}     
\begin{minipage}{1\linewidth}
\centering
\includegraphics[width=.32\textwidth]{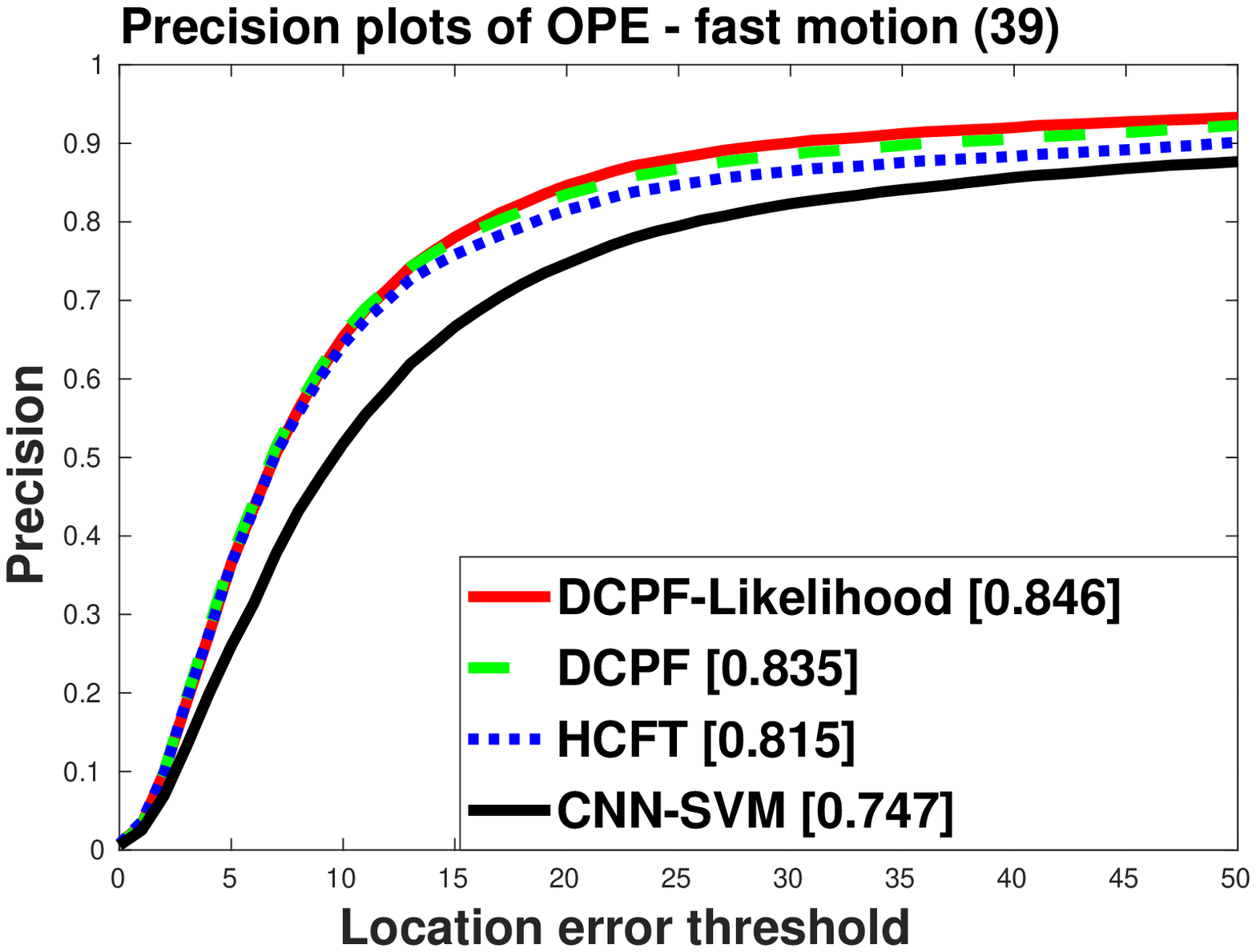}
\includegraphics[width=.32\textwidth]{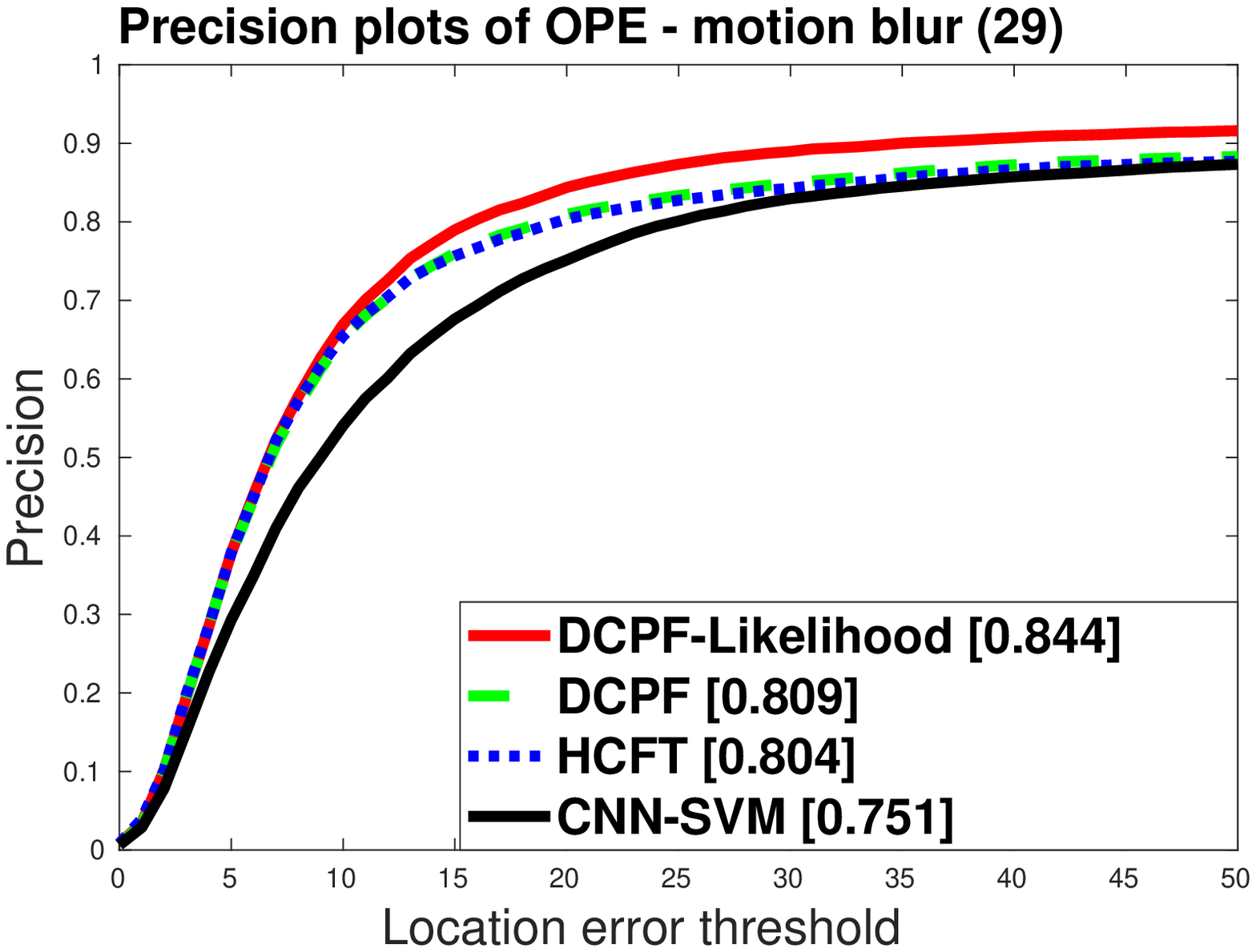}
\includegraphics[width=.32\textwidth]{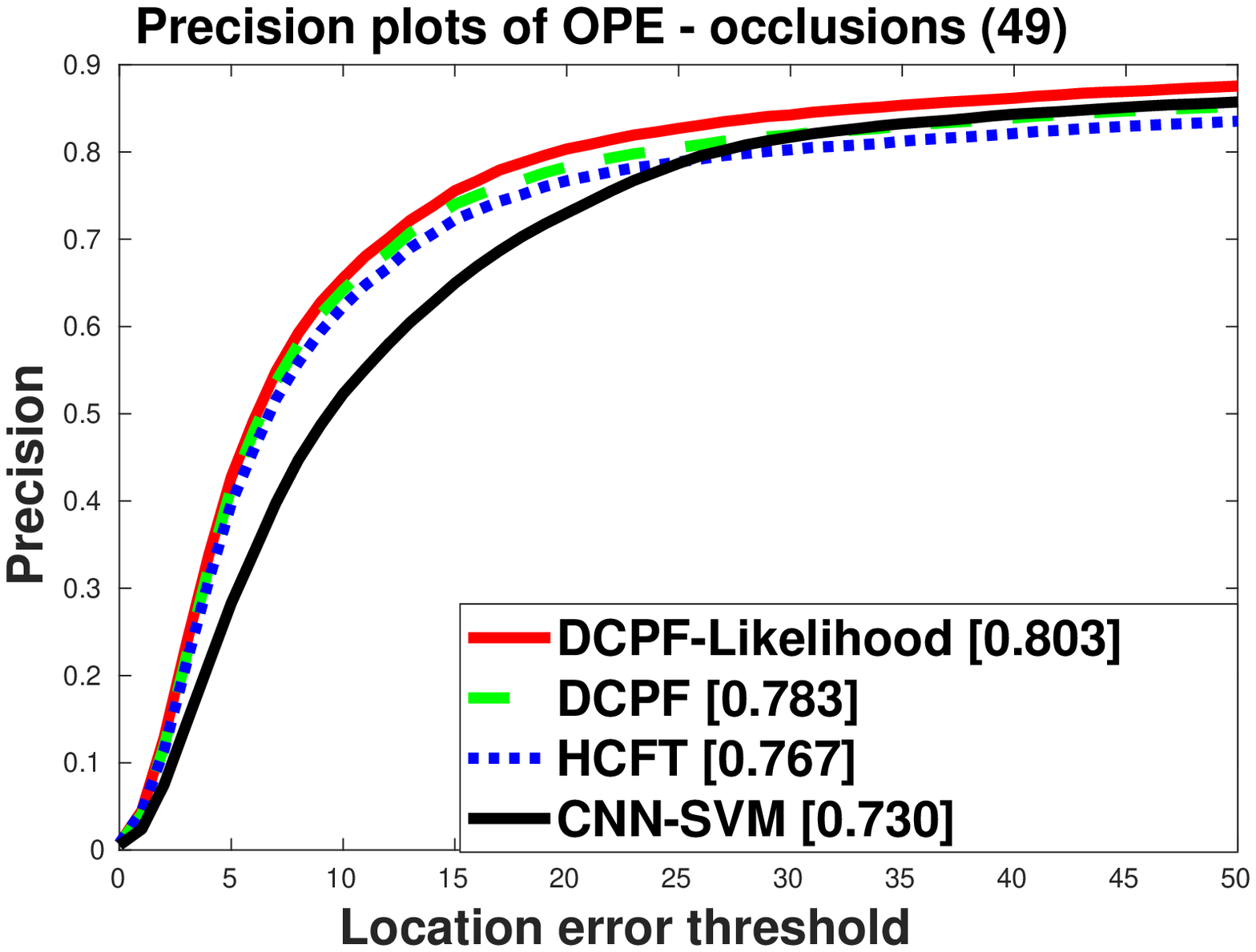}
\end{minipage}   
\begin{minipage}{1\linewidth}
\centering
\includegraphics[width=.32\textwidth]{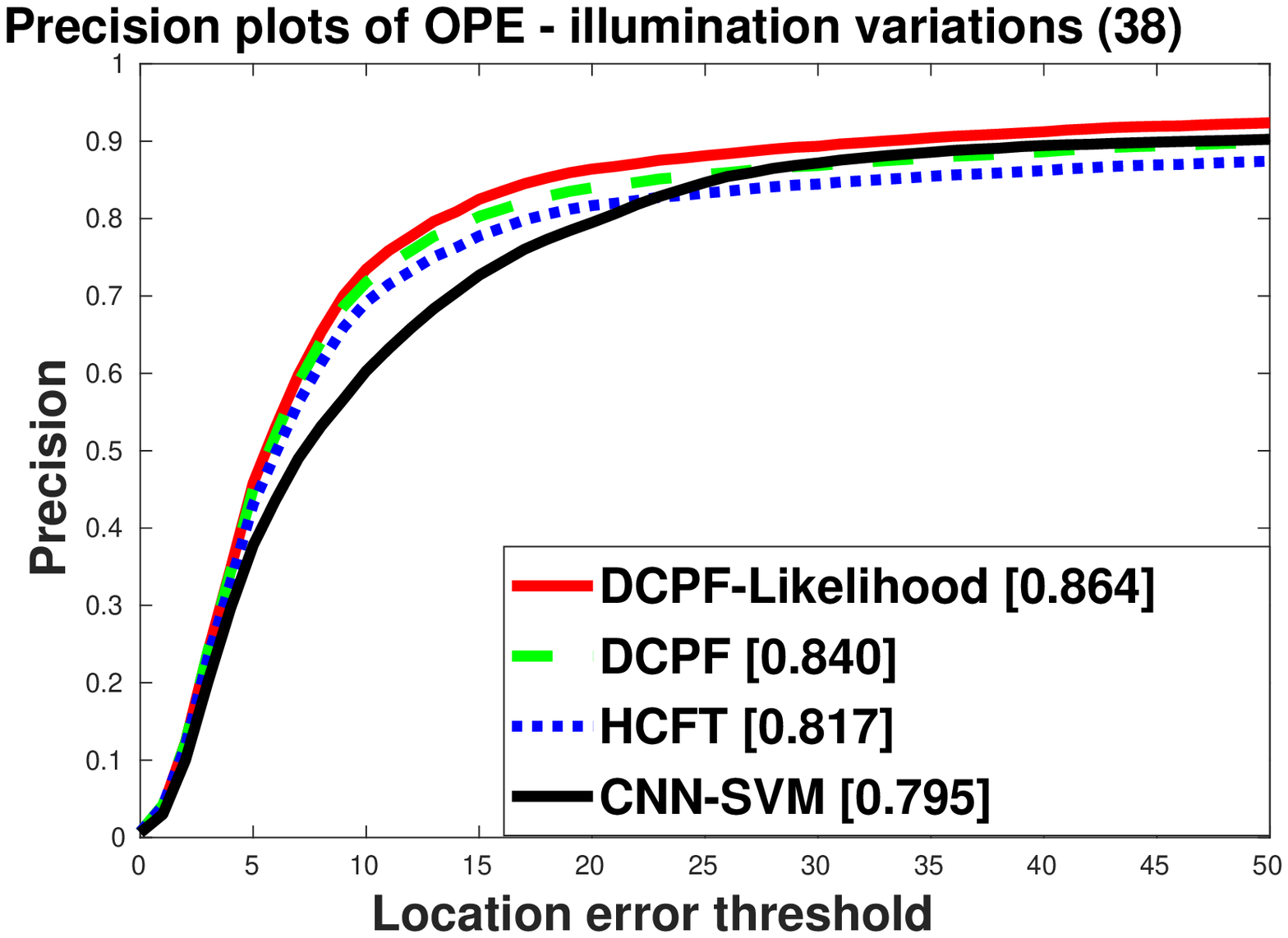}
\includegraphics[width=.32\textwidth]{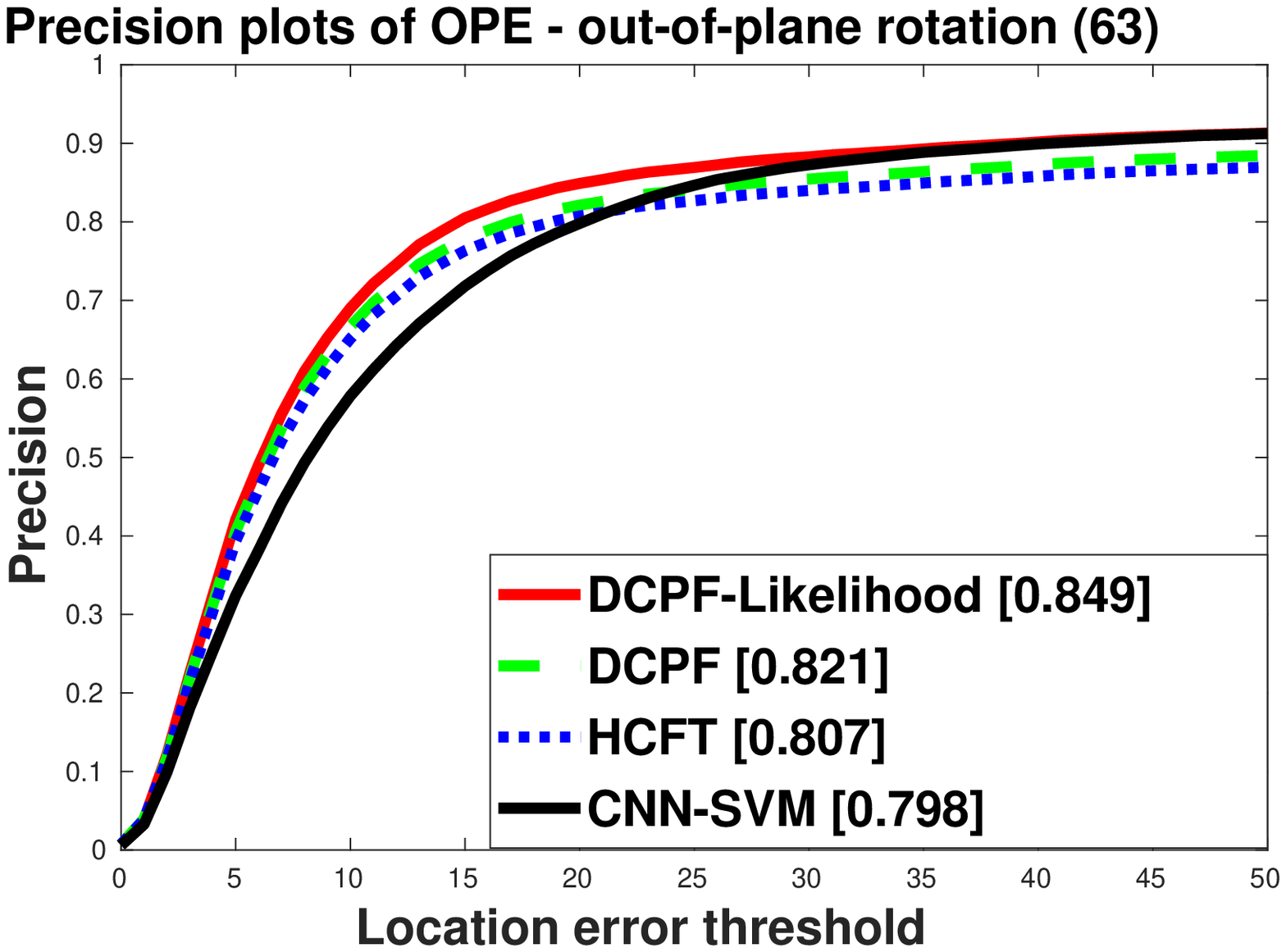}
\includegraphics[width=.32\textwidth]{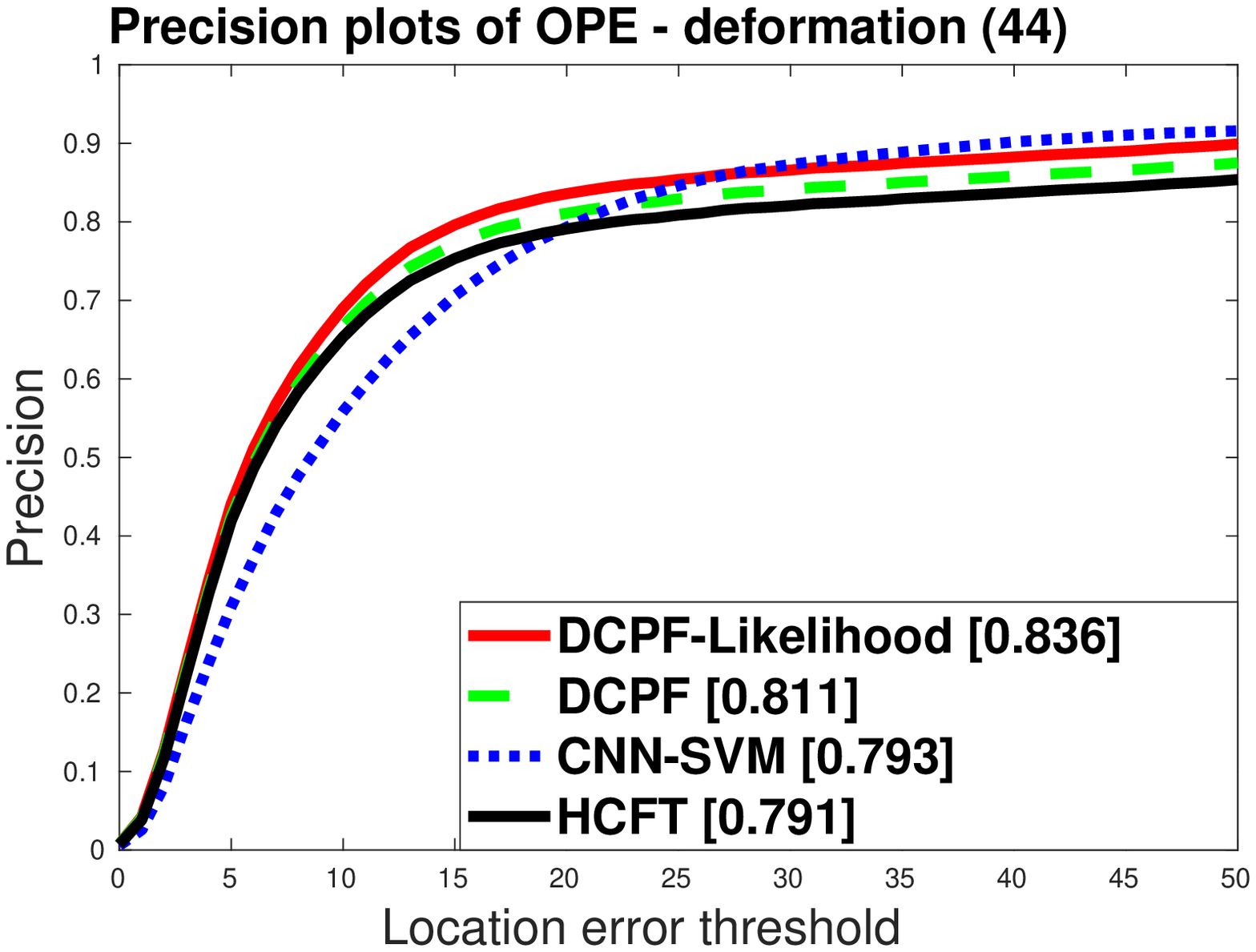}
\begin{small}
\caption{One pass evaluation of our tracker in comparison with three state-of-the-art approaches.}
\label{fig:OPE2}
\end{small}
\end{minipage}   
\end{figure*}

Fig. \ref{fig:OPE2} shows the OPE results for our tracker in comparison with DCPF, HCFT, and CNN-SVM. Our overall performance improvements over DCPF, the second best tracker, in terms of precision and success rates are $2.5\%$ and $2\%$, respectively. Our method outperforms DCPF particularly in scenarios involving occlusions (+3\%) and background clutter (+4.5\%). DCPF uses the transition distribution as the proposal density, a common approach in particle-correlation trackers. Our results show that the likelihood is a more effective proposal distribution. In scenarios involving motion blur and fast motion, our performance improvements over DCPF are around $4.5\%$ and $2\%$, respectively, because our tracker increases the variance of the likelihood distribution to spread out particles across a wider area. Our method also outperforms DCPF in scenarios involving illumination variation (+3\%), out-of-plane rotation (+3.5\%), and deformation (+3\%). Our method also decreases the computational cost of the algorithm. Our tracker uses $100$ particles, which is significantly less than the $300$ particles used in DCPF. 

\section{Conclusion} %
In this work, we propose the DCPF-Likelihood visual tracker. Our method estimates a likelihood distribution as the proposal density for a particle filter based on correlation response maps. Correlation response maps provide an initial estimate of the target location, which results in more accurate particles. Furthermore, the resulting likelihood distribution has a wider variance in challenging scenarios such as fast motion and motion blur. Our particle filter also generates a likelihood distribution for each correlation map cluster in difficult scenarios such as target occlusions. Our results on the OTB100 dataset show that our proposed visual tracker outperforms state-of-the-art methods.

\bibliographystyle{spmpsci}
\bibliography{author}

\begin{thebibliography}{10}
\providecommand{\url}[1]{{#1}}
\providecommand{\urlprefix}{URL }
\expandafter\ifx\csname urlstyle\endcsname\relax
  \providecommand{\doi}[1]{DOI~\discretionary{}{}{}#1}\else
  \providecommand{\doi}{DOI~\discretionary{}{}{}\begingroup
  \urlstyle{rm}\Url}\fi

\bibitem{tutorial}
Arulampalam, M.S., Maskell, S., Gordon, N., Clapp, T.: A tutorial on particle
  filters for online nonlinear/non-gaussian bayesian tracking.
\newblock IEEE Transactions on Signal Processing \textbf{50}(2), 174--188
  (2002)

\bibitem{dai2019visual}
Dai, K., Wang, D., Lu, H., Sun, C., Li, J.: Visual tracking via adaptive
  spatially-regularized correlation filters.
\newblock In: Proceedings of the IEEE Conference on Computer Vision and Pattern
  Recognition, pp. 4670--4679 (2019)

\bibitem{777}
Henriques, J.F., Caseiro, R., Martins, P., Batista, J.: High-speed tracking
  with kernelized correlation filters.
\newblock IEEE Transactions on Pattern Analysis and Machine Intelligence
  \textbf{37}(3), 583--596 (2015)

\bibitem{hong2015tracking}
Hong, S., You, T., Kwak, S., Han, B.: Online tracking by learning
  discriminative saliency map with convolutional neural network.
\newblock In: 32nd International Conference on Machine Learning (2015)

\bibitem{mixture}
Kawabata, T.: Multiple subunit fitting into a low-resolution density map of a
  macromolecular complex using a gaussian mixture model.
\newblock Biophysical Journal \textbf{95}(10), 4643--4658 (2008)

\bibitem{NIPS2012}
Krizhevsky, A., Sutskever, I., Hinton, G.E.: Imagenet classification with deep
  convolutional neural networks.
\newblock In: F.~Pereira, C.J.C. Burges, L.~Bottou, K.Q. Weinberger (eds.)
  Advances in Neural Information Processing Systems 25, pp. 1097--1105 (2012)

\bibitem{77}
Ma, C., Huang, J.B., Yang, X., Yang, M.H.: Hierarchical convolutional features
  for visual tracking.
\newblock In: IEEE International Conference on Computer Vision (ICCV) (2015)

\bibitem{mozhdehideep}
Mozhdehi, R.J., Medeiros, H.: Deep convolutional particle filter for visual
  tracking.
\newblock In: 24th IEEE International Conference on Image Processing (ICIP)
  (2017)

\bibitem{mozhdehideep3}
Mozhdehi, R.J., Reznichenko, Y., Siddique, A., Medeiros, H.: Convolutional
  adaptive particle filter with multiple models for visual tracking.
\newblock In: 13th International Symposium on Visual Computing (ISVC) (2018)

\bibitem{mozhdehideep2}
Mozhdehi, R.J., Reznichenko, Y., Siddique, A., Medeiros, H.: Deep convolutional
  particle filter with adaptive correlation maps for visual tracking.
\newblock In: 25th IEEE International Conference on Image Processing (ICIP)
  (2018)

\bibitem{qi2016hedged}
Qi, Y., Zhang, S., Qin, L., Yao, H., Huang, Q., Lim, J., Yang, M.H.: Hedged
  deep tracking.
\newblock In: IEEE Conference on Computer Vision and Pattern Recognition
  (CVPR), pp. 4303--4311 (2016).
\newblock \doi{10.1109/CVPR.2016.466}

\bibitem{234}
Simonyan, K., Zisserman, A.: Very deep convolutional networks for large-scale
  image recognition.
\newblock In: International Conference on Learning Representations (ICLR)
  (2015)

\bibitem{WuLimYang}
Wu, Y., Lim, J., Yang, M.H.: Online object tracking: A benchmark.
\newblock In: IEEE Conference on Computer Vision and Pattern Recognition (CVPR)
  (2013)

\bibitem{Redetection}
Yuan, D., Lu, X., Liang, Y., Zhang, X.: Particle filter re-detection for visual
  tracking via correlation filters.
\newblock Multimedia Tools and Applications  (2018)

\bibitem{zhang2018visual}
Zhang, M., Wang, Q., Xing, J., Gao, J., Peng, P., Hu, W., Maybank, S.: Visual
  tracking via spatially aligned correlation filters network.
\newblock In: Proceedings of the European Conference on Computer Vision (ECCV),
  pp. 469--485 (2018)

\bibitem{CPF}
Zhang, T., Liu, S., Xu, C.: Correlation particle filter for visual tracking.
\newblock IEEE Transactions on Image Processing \textbf{27}(6), 2676--2687
  (2018)

\end{thebibliography}

\end{document}